# On the focusing of thermal images

Marcos Faundez-Zanuy (1), Jiří Mekyska (2), Virginia Espinosa-Duró (1)

(1) EUP Mataró, Pl. Santi Ortego, Tecnocampus, 08302 MATARO (BARCELONA) SPAIN

(2) Department of Telecommunications, Faculty of Electrical Engineering and Communication, Brno University of Technology, Purkynova 118, 612 00 Brno, Czech Republic

faundez@eupmt.es, xmekys01@stud.feec.vutbr.cz espinosa@eupmt.es,

**Abstract** – In this paper we present a new thermographic image database suitable for the analysis of automatic focus measures. This database consists of 8 different sets of scenes, where each scene contains one image for 96 different focus positions. Using this database we evaluate the usefulness of six focus measures with the goal to determine the optimal focus position. Experimental results reveal that an accurate automatic detection of optimal focus position is possible, even with a low computational burden. We also present an acquisition tool able to help the acquisition of thermal images. To the best of our knowledge, this is the first study about automatic focus of thermal images.

## 1. INTRODUCTION

Image acquisition is a frequent task in a large series of applications, including entertainment, medicine, industry, security, weather analysis, etc. In all these cases the outcome of the camera focus determines the quality of the image. Several approaches exist for focusing an image. The first classification is between manual and automatic. In the former case, the user must trim the optic till he obtains a satisfactory result. This trim can be manual or helped with an engine that moves the focus. In both cases, the user must decide the optimal focus position. This process is usually tedious and can be complicated for several users with sight problems (such as myopia), lack of skill, etc. In addition, typical digital camera visors do not provide enough resolution to fully appreciate if the image is focused or blurred. However, when several objects exist at different focal planes, the user can select which objects appear focused and which ones blurred. This is sometimes difficult with automatic focus.

When using an automatic system, there is an electronic system that performs the task without the user interaction. Automatic focus can be split into two main categories. Active systems require the presence of an integrated sensor. Normally, this sensing equipment will emit an indicating signal such as ultrasound wave or infrared ray and then wait for its reflection. The distance between object of interest and imaging equipment is then estimated based on the transmission duration between the emitted signal and its reflected counterpart. Based on this estimated distance, the camera parameters such as the lens position and aperture setting are adjusted accordingly in order to attain the best focused position. Passive systems perform a set of focus measurements on the acquired image. This can be done in spatial or frequency domain. The spatial techniques require lesser number of operations and are more suitable for real time applications. In frequency domain the procedure is quite straight forward because focused images contain sharp edges, associated to high frequency content. Table 1 summarizes the existing kinds of focus techniques.

While there is a considerable amount of work done on visible images [1-8], we are not aware about studies on thermal images.

| Focus | kind | Domain | Pros | Cons |
|---|---|---|---|---|
| Manual | Manual/ motorized | Spatial | • Cheap<br>• User can select the focused object inside the scene. | • tedious<br>• complicated for several users |
| automatic | Active | Temporal | • Can work in complete darkness | • Requires external sensor.<br>• Some objects tend to absorb the transmitted signal energy.<br>• There is distance-to-subject limitation (6 meters).<br>• A source of infrared light from an open flame (birthday cake candles, for instance) can confuse the infrared sensor.<br>• may fail to focus a subject that is very close to the camera. |
| | Passive | Spatial | • Suitable for real time<br>• There is no distance-to-subject limitation | • Requires good illumination and image contrast. |
| | | Frequency | • High accuracy<br>• There is no distance-to-subject limitation | • Requires good illumination and image contrast.<br>• High computational complexity |

Table 1. Summary of kinds of focus

### 1.1 Visible and thermal images

The portion of the electromagnetic spectrum visible for a human eye roughly ranges from 300nm to 700nm when measured in terms of daylight conditions, being not a flat response and showing a maximum sensitivity at 555nm. This is also called photopic curve and matches with the CIE (Commission Internationale de l'Éclairage) standard curve used in the CIE 1931 color space. In addition, this curve shifts itself towards shorter wavelengths zone, in darkness conditions becoming the scotopic curve which presents a new peak sensitivity at 510nm. The near infrared (NIR) windows lie just out of human response window, and the Medium Infrared (MIR) and Far IR (FIR) are far beyond the human response region.

An especially interesting sub-band of FIR spectrum lies from 3 μm to 14 μm, called Thermal Infrared (TIR), which humans experience every day in the form of heat or thermal radiation. This special band of the spectrum presents two important windows called Mid Wave IR (MWIR) comprised in the range between 3 and 5 μm, and Long Wave IR (LWIR) that lies in the range from 8 μm to14 μm. Between them exist a blocked band due to the contamination by solar reflectance and water vapor absorption. Figure 1 shows the electromagnetic spectrum paying special attention to the overall existing infrared sub-bands presented above.

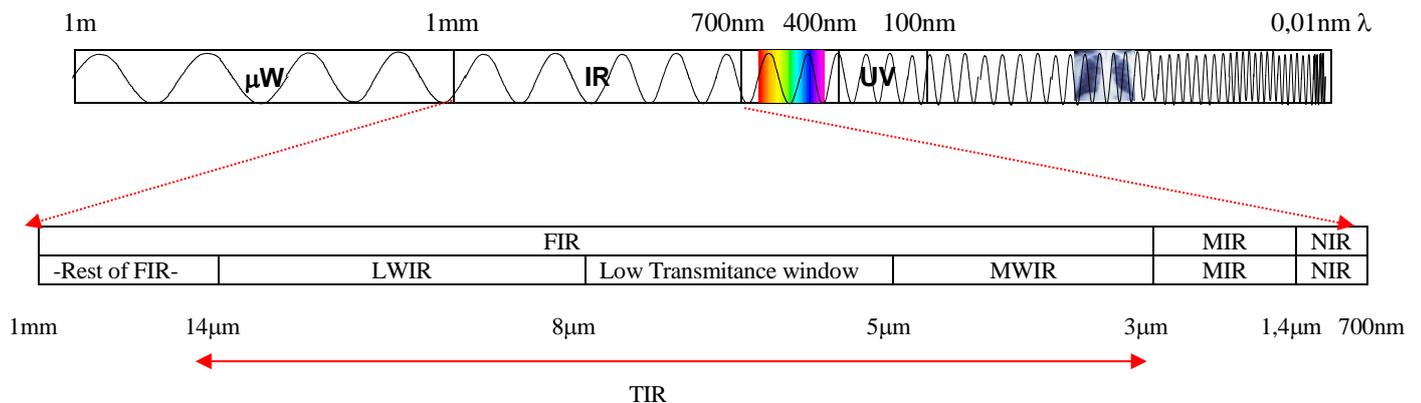

Fig. 1. Electromagnetic spectrum.

Since EM radiation, most strongly infrared one, is emitted by all objects based on their temperatures, according to the black body radiation law, it is possible to "see" live beings and warm objects with or without visible illumination. This overall kind of images can be acquired by means of a thermographic camera. Is also important to note that exist two kind of thermal imaging cameras depending of the type of incorporated sensor: Uncooled ones, equipped with uncooled sensors, specially designed to work in the LWIR band, where terrestrial targets radiate most of their infrared energy and uncooled detection is easy and cooled ones, equipped with cooled sensors to cryogenic temperatures which are highly most sensitive to discriminate small temperature differences in scene temperature and are generally designed to acquire images in both, MWIR or LWIR bands.

In [9-10] we studied the redundancy between several spectral bands using information theory and we showed that there is complementary information between different wavelengths. This emphasizes the interest for thermographic imaging applications.

Independently of the wavelength, the focus problem exists for thermal images as well as for visible ones. For this reason it is important to dispose of objective criteria to evaluate if a given image is focus or not. This paper is devoted to the study of thermal image focus and is organized as follows: section two presents several focus measures. Section three presents the database specially acquired for this paper and the experimental results. Section four summarizes the main conclusions.

## 2. FOCUS MEASURES

In this paper we will evaluate several focus measures suitable for automatic focus. While this topic has attracted the attention of the scientific community when dealing with visible images, this is not the case with thermal images. The study of focus on thermal images is challenging due to the following main reasons:

1. It is harder to manually focus a thermal image than a visible one, because we are not used to see thermal images. This means that some skill and habituation to use thermal cameras is required.

2. Image cameras usually incorporate a small screen with little resolution. Although a human operator can consider that a given image is focused, sometimes he realizes that it is blurred when visualized in a larger screen (once upon transferred from the camera to a personal computer).

3. To the best of our knowledge, there are no available studies on this topic, and thermal cameras do not incorporate an automatic focus. In our case, the TESTO 880-3 camera incorporates an engine able to move the focus, but the trim must be performed visually (motorized manual focus described in table 1). The focus can be reached by making small adjustments of the focusing knob until edges appear sharp.

A typical focus measure should satisfy these requirements:

1. It should be independent of image content. However, if the image contains a large amount of thin details, it is easier to focus.
2. Monotonic with respect to blur. If we move away from the optimal focus position, the focus measure should decrease monotonically. Typically this will happen when moving the focus in both directions (left and right).
3. The focus measure must be unimodal, that is, it must have one and only one maximum value. While this is simple for "flat" scenes, this cannot be true for scenes with objects at different focal distances. For instance, if the nearer object if focused, the most distant will be blurred and vice versa. However, computational photography methods let to combine multi-focus images [8].
4. Large variation in value with respect to the degree of blurring. This will permit a sharp peak (maximum focus value).
5. Minimal computation complexity. For real time image acquisition the feedback about the focus measure should be obtained as quickly as possible.
6. Robust to noise: In presence of noise the maximum focus value should be stable and unique. In this aspect is important to emphasize that near infrared and visible images are sensible to illumination conditions. If illumination is not good enough, the image would be noisy. However, thermal cameras are not affected by illumination because they acquire the heat emission, not the illumination reflection.

In order to obtain the most suitable measure for a thermal image, we have performed a set of experiments with several images and focus measures. Although we have not found studies about the focus of thermal images, we think that this is due to the high price of thermographic cameras and the reduced amount of thermal images for home applications. Nevertheless a human operator considers that a thermal image is focus when it presents the highest amount of details and sharpness. In fact the operation instructions of commercial thermographic cameras assert that a thermal image is focus when edges appear sharp. Thus, focus measures should be similar to those used for visible images, which are mainly sharpness measures. Next sections describe these measures.

### 2.1 Variance
A very simple measure is the variance of the image. Blurred images have smaller variance than focused ones.

### 2.2 Energy of image gradient
The energy of image gradient (EOG) is based on the vertical and horizontal gradients of the image, and is obtained as:

$$EOG = \sum_{x=1}^{M-1} \sum_{y=1}^{N-1} (f_x^2 + f_y^2),$$

### 2.3 Tenengrad
This measure (Krotkov 1987) is based on the gradient magnitude from the Sobel operator:

$$Tenengrad = \sum_{x=2}^{M-1} \sum_{y=2}^{N-1} \big(\nabla S(x,y)\big)^2 \quad \text{for} \ \ \nabla S(x,y) > T,$$

where T is a discrimination threshold value, and $\nabla S(x,y)$ is the Sobel gradient magnitude value.

### 2.4 Energy of Laplacian of the image
Energy of Laplacian can be computed as:

$$EOL = \sum_{x=2}^{M-1} \sum_{y=2}^{N-1} (f_{xx} + f_{yy})^2,$$

where

$$f_{xx} + f_{yy} = -I(x-1,y-1) - 4I(x-1,y) - I(x-1,y+1) - 4I(x,y-1) + 20I(x,y) - 4I(x,y+1)$$
$$- I(x+1,y-1) - 4I(x+1,y) - I(x+1,y+1).$$

### 2.5 Sum-modified Laplacian
Nayar and Nakagawa [1] noted that in the case of the Laplacian the second derivatives in the x- and y-directions can have opposite signs and tend to cancel each other. Therefore, he proposed the sum modified Laplacian (SML), which can be obtained by means of:

$$SML = \sum_{i=x-W}^{x+W} \sum_{j=y-W}^{y+W} \nabla^2_{ML} f(i,j) \quad \text{for} \ \ \nabla^2_{ML} f(i,j) \geq T,$$

where $T$ is a discrimination threshold value and:
$$\nabla^2_{ML} f(x,y) = |2I(x,y) - I(x-step,y) - I(x+step,y)| + |2I(x,y) - I(x,y-step) - I(x,y+step)|.$$

In order to accommodate for possible variations in the size of texture elements, Nayar used a variable spacing (step) between the pixels to compute ML. The parameter $W$ determines the window size used to compute the focus measure.

### 2.6 Crete et al.

To be independent from any edge detector and to be able to predict any type of blur annoyance, Crete et al. [2] proposed an approach, which is not based on transient characteristics but on the discrimination between different levels of blur perceptible on the same picture. The algorithm for the no-reference blur measurement can be described by these formulas:

$$h_v = \frac{1}{9}[1\ 1\ 1\ 1\ 1\ 1\ 1\ 1\ 1],$$

$$h_h = \frac{1}{9}[1\ 1\ 1\ 1\ 1\ 1\ 1\ 1\ 1]^T,$$

$$b_v = h_v * I(x, y),$$

$$b_h = h_h * I(x, y),$$

Where $h_v$ and $h_h$ are the impulse responses of horizontal and vertical low-pass filters which are used to make the blurred version of the image $I(x, y)$. In the next step the absolute difference images $Di_v(x, y), Di_h(x, y), Db_v(x, y)$ and $Db_h(x, y)$:

$$Di_v(x, y) = \text{abs}\big(I(x, y) - I(x - 1, y)\big) \text{ for } x = 1, 2, \ldots, M - 1, y = 0, 1, \ldots, N - 1,$$

$$Di_h(x, y) = \text{abs}\big(I(x, y) - I(x, y - 1)\big) \text{ for } x = 1, 2, \ldots, N - 1, y = 0, 1, \ldots, M - 1,$$

$$Db_v(x, y) = \text{abs}\big(b_v(x, y) - b_v(x - 1, y)\big) \text{ for } x = 1, 2, \ldots, M - 1, y = 0, 1, \ldots, N - 1,$$

$$Db_h(x, y) = \text{abs}\big(b_v(x, y) - b_v(x, y - 1)\big) \text{ for } x = 1, 2, \ldots, N - 1, y = 0, 1, \ldots, M - 1.$$

Then the variation $V_v$ and $V_h$ of neighboring pixels is analyzed:

$$V_v = \max\big(0, Di_v(x, y) - Db_v(x, y)\big) \text{ for } x = 1, 2, \ldots, M - 1, y = 0, 1, \ldots, N - 1,$$

$$V_h = \max\big(0, Di_h(x, y) - Db_h(x, y)\big) \text{ for } x = 1, 2, \ldots, M - 1, y = 0, 1, \ldots, N - 1.$$

If the variation is high, then the initial image was sharp on the other hand the initial image $I(x, y)$ was already blurred. In the next step the sum of coefficients $Di_v(x, y), Di_h(x, y), V_v(x, y)$ and $V_h(x, y)$ is calculated in order to compare the variations from the initial picture:

$$Si_v = \sum_{x, y = 1}^{M - 1, N - 1} Di_v(x, y),$$

$$Si_h = \sum_{x, y = 1}^{M - 1, N - 1} Di_h(x, y),$$

$$SV_v = \sum_{x, y = 1}^{M - 1, N - 1} V_v(x, y),$$

$$SV_h = \sum_{x, y = 1}^{M - 1, N - 1} V_h(x, y).$$

The vertical $Bi_v$ and horizontal $Bi_h$ blur values from range 0 to 1 are calculated according to equations:

$$Bi_v = \frac{Si_v - SV_v}{Si_v},$$

$$Bi_h = \frac{Si_h - SV_h}{Si_h}.$$

The algorithm is designed to calculate the blur value, but for our purpose the value describing the sharpness is more useful. This value $S$ can be obtained easily according to formula:

$$S = 1 - \max(Bi_v, Bi_h).$$

It means that the sharp images will have the value $S$ closer to 1 and the blurred images closer to 0.

## 3. EXPERIMENTAL RESULTS

In order to analyze the focus of a thermal image using a focus measure we have acquired several databases. Using these databases, we have evaluated the focus measure for each image and plotted against the focus position.

### 3.1 Database acquisition procedure

In order to perform the experiments, we have acquired a special purpose database using a thermographic camera TESTO 880-3. This camera is equipped with an uncooled detector and has a spectral sensitivity range from 8 to 14µm. It has a removable Germanic optic lens, and an approximate cost of 8.000 EUR. It provides the following main features:

- Image resolution: 160×120 pixels
- Optical field/min. focus distance: 32° x 24° / 0,1 m
- Thermal sensitivity (NETD) <0,1 °C at 30 °C
- Geometric resolution: 3,5 mrad
- Detector type: FPA 160 x 120 pixels, a.Si, temperature-stabilized

The database consists of several image sets. In each set, the camera acquires one image of the scene at each lens position. In our case we have manually moved the lens in a 1mm steps and this provides a total amount of 96 positions. Thus, each set consists of 96 different images of the same scene. For this purpose we have attached a milimetric tape to the objective, and we have used a stable tripod in order to acquire always the same scene for each scene position.

We have developed a program able to control the thermographic camera from a laptop. This program shows the focus measure values in order to facilitate the image acquisition. The image is stored in *.bmp file format.This program will be freely available after the publication of the paper, as well as the database.

## 3.2 Database description

We tried to acquire different kind of images according to amount of details (information), as well as different deepness of the image. It should be easier-to-focus an image with large amount of details, because a blurred image will be more evident. In addition, in similar way to visible images, it should be more difficult to focus an image with several objects, when each one is located at a different focal distance. In addition, we looked for static scenes, because the whole sequence of images should contain the same objects (same position and temperature).

We have collected five different databases:

a)  Telematic equipment (TE): it consists of a four set of images of a same scene (telematic equipment). Set one (TE1) is acquired with a one meter distance from the scene to the camera. Set two (TE2) is acquired at a distance of two meters, set three (TE3) at three meters and set four (TE4) at four meters. Figure 2 shows the most focused image of each set. We want to point out that, when moving the camera away from the scene, more objects appear on it. On the other hand, this database contains a scene that can be considered to be contained in a flat plane. Thus, it is acquiring mainly a two dimensional object with almost no deepness.

b)  Electronic circuit (EC): it consists of a single set of images of a same scene (electronic circuit with components at different temperature and deepness). Figure 3 on the upper left shows the most focused image of this database. It is important to emphasize that in this case we are acquiring a very near object, which has a considerable deepness. Thus, it is not possible to focus all the objects simultaneously.

c)  Laptop transformer (LT): It consists of a single set of images of a same scene (the transformer of a laptop computer). Figure 3 on the upper right shows the most focused image of this database. Figure 4 shows several images at different focus positions. The image in the middle is focused. The other ones are blurred.

d)  Corridor and fluorescents (CF): it consists of a single set of images of a same scene (a corridor at the university, with several fluorescents on the ceiling). Figure 3 on the bottom left shows the most focused image of this database.

e)  Heater (H): it consists of a single set of images of a heater. It is important to emphasize that this scene contains a large amount of detail because the metallic parts are warmer than the space in the middle. Figure 3 on the bottom right shows the most focused image of this database.

This new database specially acquired for this work will be distributed for free for the scientific community. The whole database consists of $8 \times 96 = 768$ images.

Experimental results revealed that the variance was not reliable with some image sets and it has been discarded as a focus measure.

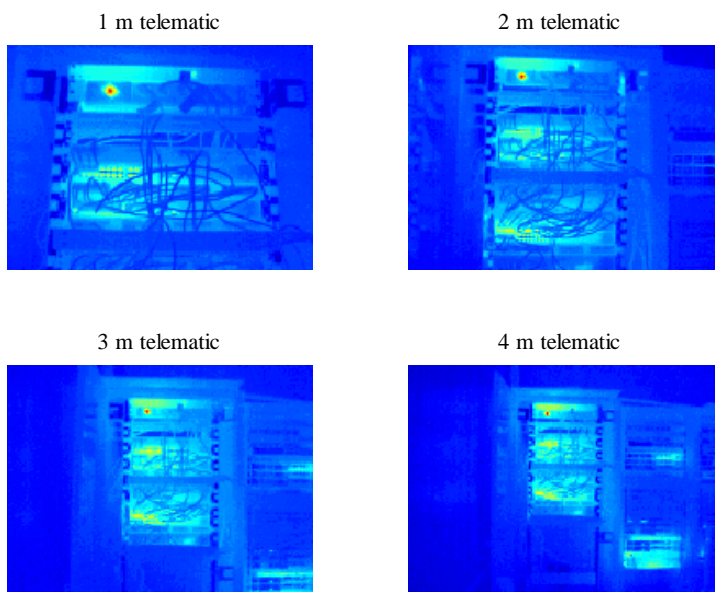

Figure 2. Best image of each set for the telematic equipment database, at the four evaluated distances.

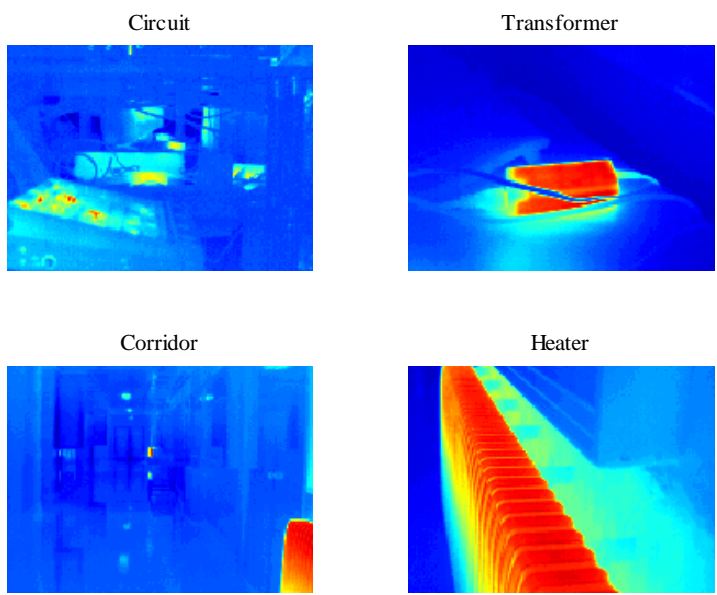

Figure 3. Best image of Electronic circuit, Laptop transformer, corridor and fluorescent, and heater databases.

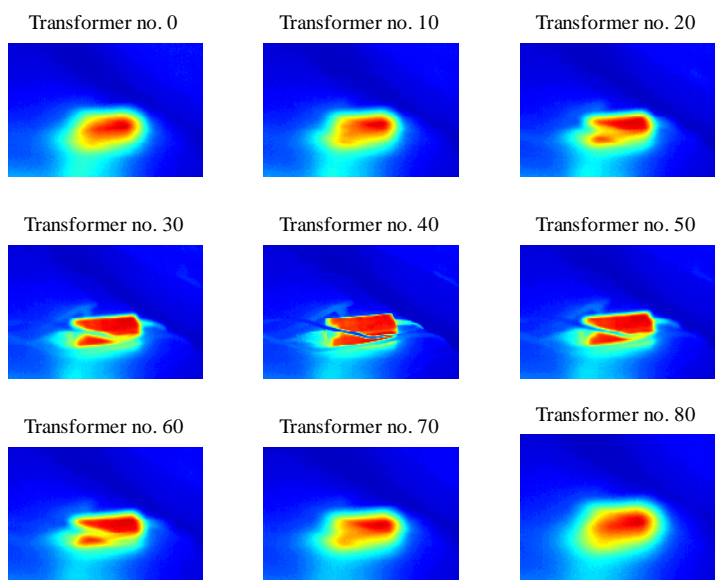

Figure 4. Example of different focus positions of a laptop transformer.

### 3.3 Experimental results

In this section we present the values of the different focus measures described in section 2. Figure 5 shows the focus values described in sections 2.2 to 2.7 using the TE database. It can be observed that the curve shape is mainly the same but there is, as expected, a shift to the left when the distance is higher. This is due to the change of focal distance. In addition, Tenengrad method fails to provide an accurate peak, which implies that it is not possible to find an accurate focus using this measure.

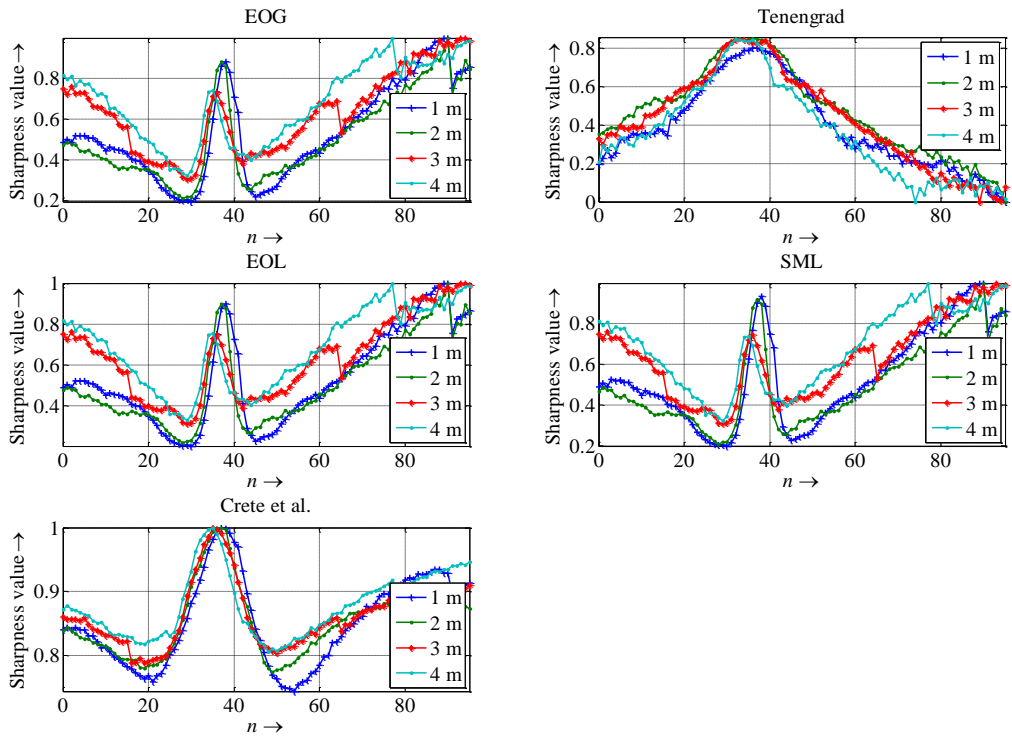

Figure 5. Focus measures using the TE database.

Figure 6 shows the experimental results for the EC database. Similar results are obtained in figure 7 with LT database, but in this case the peak is again sharper due to the simplicity of the scene. Figure 8 shows the experimental results for the EOG, Tenengrad, EOL, SML and Crete et al. algorithms respectively using the CF database. It can be observed that most of the methods agree in a sharp peak for optimal focus position, which has been validated by human inspection. On the other hand it is also clear that Tenengrad and Crete et al. algorithms fail to provide a reliable estimation. In figure 8 we obtain similar results than figure 6 but the focus peak is more evident. This is due to the fact that EC scene is harder to focus due to the image deepness. Again, Tenengrad and Crete et al. fail to provide a reliable focus position.

Figure 9 presents the experimental results with the H database. In this case, all the methods provide the same focus position with the exception of the Tenengrad algorithm.

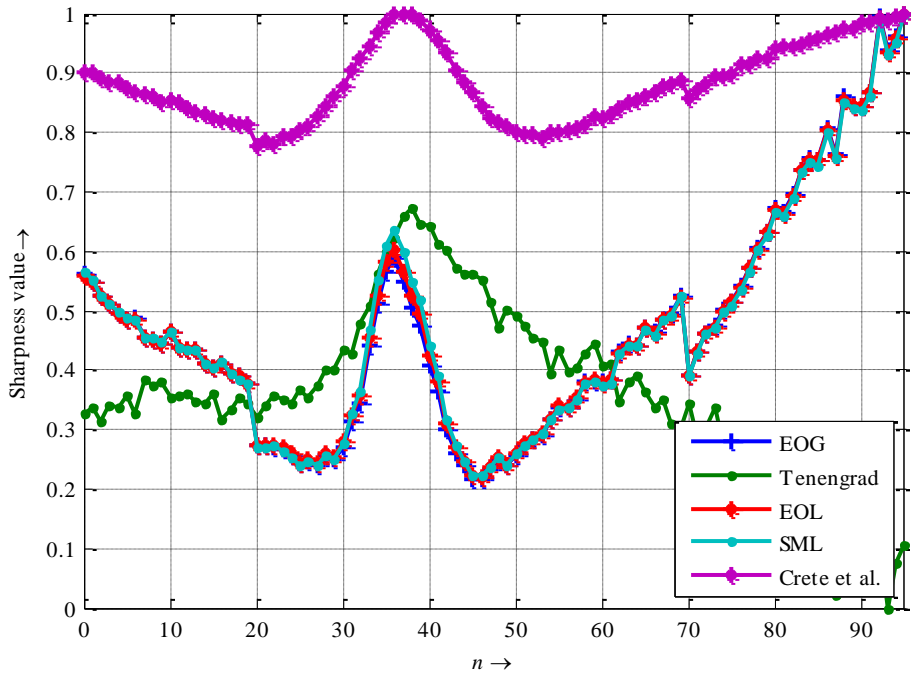

Figure 6. Focus measures using the EC database.

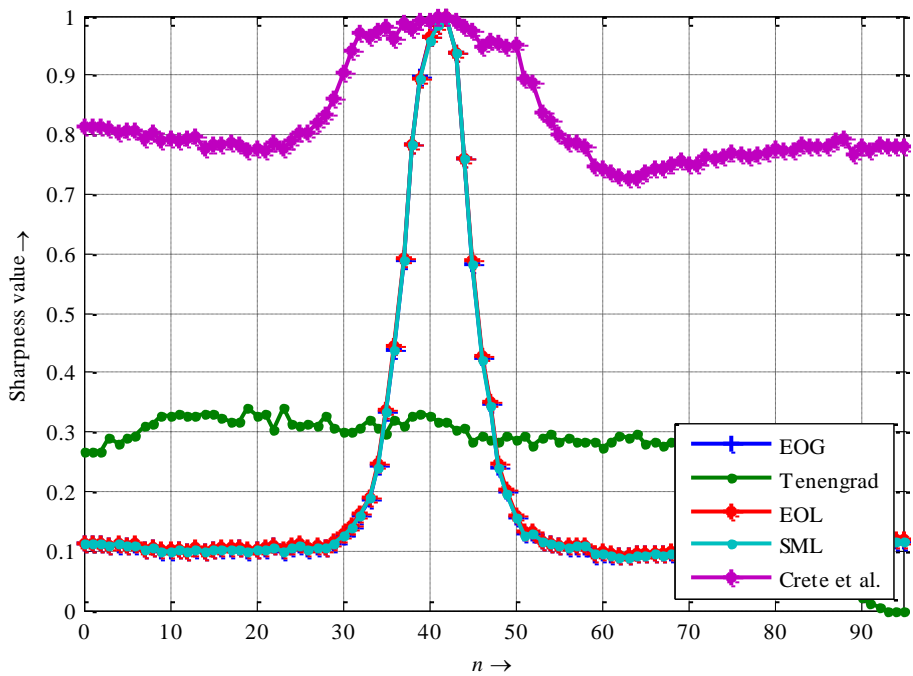

Figure 7. Focus measures using the LT database.

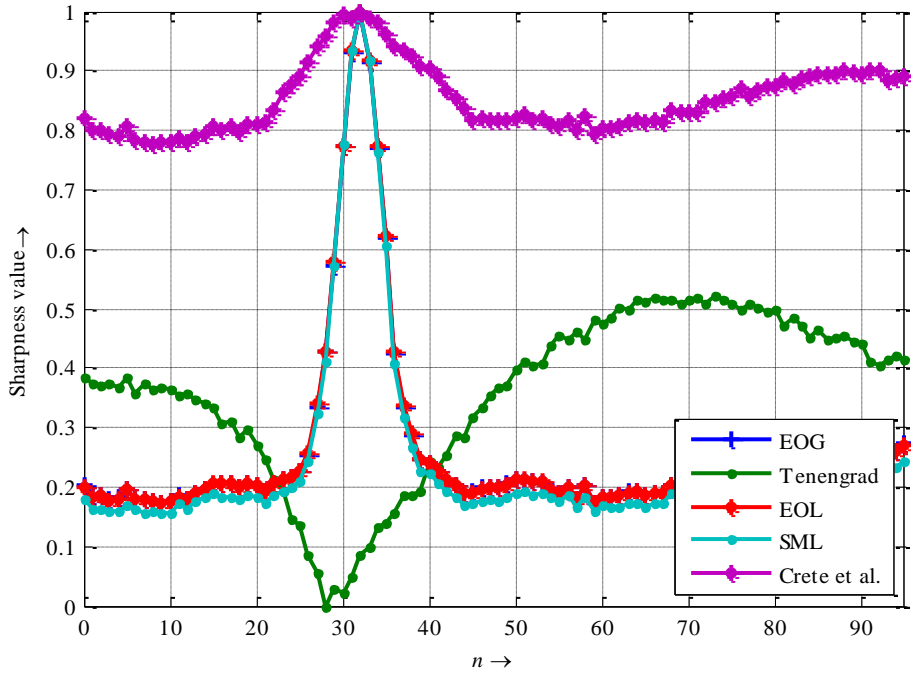

Figure 8. Focus measures using the CF database.

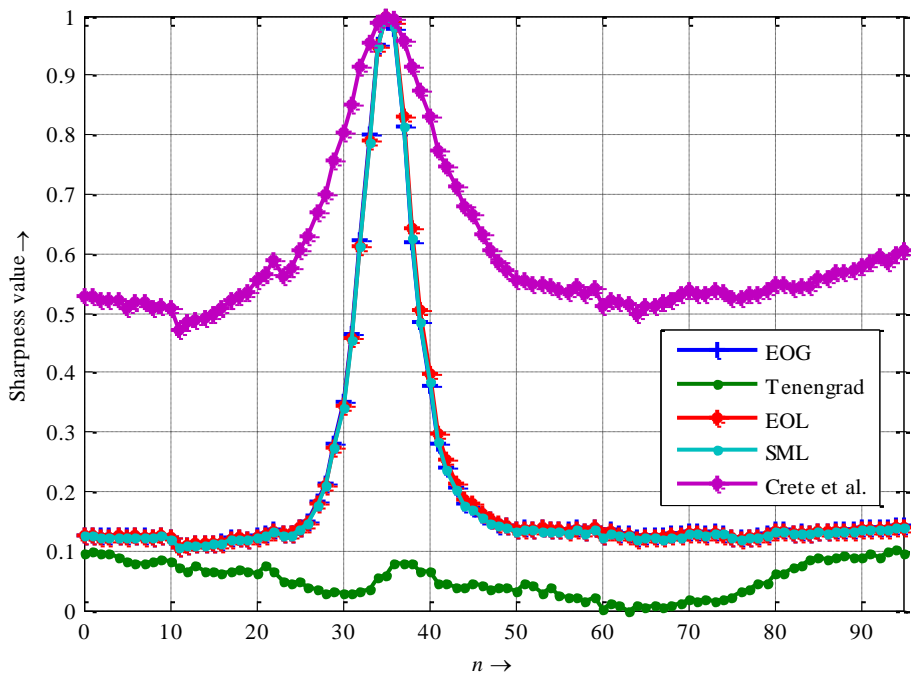

Figure 9. Focus measures using the H database.

Another useful evaluation consists of the computational time required to evaluate one image of a given database for each algorithm. Computational time has been obtained with optimized (we have avoided loops) MATLAB algorithms. This was tested on 32-bit version 7.4.0.287 (R2007a). Matlab was driven on laptop with processor Intel Core 2 Duo 2.4 GHz, 4 GB RAM and operating system Microsoft Windows Vista Home Premium. A desirable property is a low computational time, because this algorithm can be integrated in an automatic passive focus system (see table 1). Table 2 summarizes the computational time of each algorithm. As can be observed, all of them are reasonably fast. The EC set requires more computational time because the images contain more details that the other sets.

| Database | Method | | | | |
|---|---|---|---|---|---|
| | EOG | Tenengrad | EOL | SML | Crete et al. |
| TE 1 | 0,70 | 2,55 | 1,56 | 1,16 | 2,50 |
| TE 2 | 0,70 | 2,55 | 1,56 | 1,14 | 2,46 |
| TE 3 | 0,70 | 2,54 | 1,57 | 1,15 | 2,45 |
| TE 4 | 0,70 | 2,54 | 1,56 | 1,14 | 2,47 |
| EC | 0,70 | 2,52 | 1,56 | 1,14 | 2,46 |
| LT | 0,70 | 2,53 | 1,56 | 1,18 | 2,47 |
| CF | 0,80 | 2,52 | 1,59 | 1,20 | 3,35 |
| H | 0,70 | 2,53 | 1,56 | 1,17 | 2,46 |
| Maximum value | 0,80 | 2,55 | 1,59 | 1,20 | 3,35 |

Table 2. Computational time (ms) for each method

### 4. CONCLUSIONS

In this paper we have discussed the automatic focus problem. We have applied it, for the first time, with thermal images. For this purpose, we have acquired a new thermal database which consists of 8 different scenes, acquiring a set of 96 snapshots for each scene. These scenes have been selected in order to include a large amount of variability (low/high amount of details, low/high deepness, etc.). This database will be freely distributed at no cost for the scientific community, as well as a software acquisition tool that indicates the focus measure of a given thermal image.

We have presented a set of measurements and we have found the following conclusion:

- It is possible to automatically focus a thermal image: some operators can find an optimum focus position that matches the human decision. Among these measures, only energy of image Gradient (EOG), Energy of Laplacian of the image (EOL) and sum-modified Laplacian (SML) offer good performance in all the scenarios. These measures provide an accurate and sharp peak which clearly identifies the optimal focus position.
- We have observed that Tenengrad operator was unable to provide an accurate peak in some image sets. Especially those with low thermal details, such as LT and CF. We do not want to conclude that this algorithm fails when applied to thermal images. We think that this is motivated by the low resolution of the thermal images (160x120). This operator generates a blurred image, and probably this blurred image, when calculated on low resolution images with low amount of details, cannot produce enough good results due to the lack of information.
- In general, the simplest operators according to computational burden provide the best experimental results. Probably the more sophisticated algorithms require more statistical information (more pixels) in order to provide better results.
- Considering computational issues, EOG performs the analysis in less than 0.8 ms, when programmed in MATLAB. This implies that this operator is suitable for a fast automatic focus.
- Using the TE database and increasing the distance from the camera to the object, we can be observed that the curve shape is mainly the same. There is a shift to the left, as expected, when the distance is higher. This is due to the change of focal distance.
- Although we made an effort to acquire different kind of scenes, we have mainly obtained the same performance with all of them. We could not obtain the typical curve with several peaks typical of images with several objects at different focal distances. We think that this is mainly due to the low resolution images obtained with a thermal camera. While visible cameras can offer resolutions of several million pixels, thermal cameras are typically below one million pixels.

### 5. ACKNOWLEDGEMENTS

This work has been supported by FEDER and MEC, TEC2009-14123-C04-04. We also want to acknowledge the COST OC08057 project, KONTAKT-ME 10123 (Research of Digital Image and Image Sequence Processing Algorithms), project SIX (CZ.1.05/2.1.00/03.0072) and project VG20102014033.